\documentclass[acmsmall]{acmart}

\AtBeginDocument{%
  \providecommand\BibTeX{{%
    \normalfont B\kern-0.5em{\scshape i\kern-0.25em b}\kern-0.8em\TeX}}}

\setcopyright{acmcopyright}
\copyrightyear{2021}
\acmYear{2021}
\acmDOI{00.0001/0000001.0000001}
\acmJournal{TOMM}

\usepackage{amsmath}
\usepackage{algorithmic}
\usepackage{array}
\usepackage{float}
\usepackage{CJKutf8}
\usepackage{color}
\usepackage{amsfonts}
\usepackage{booktabs}
\usepackage{multirow}
\usepackage{multicol}
\usepackage{threeparttable}

\usepackage{array}
\newcommand{\PreserveBackslash}[1]{\let\temp=\\#1\let\\=\temp}
\newcolumntype{C}[1]{>{\PreserveBackslash\centering}p{#1}}
\newcolumntype{R}[1]{>{\PreserveBackslash\raggedleft}p{#1}}
\newcolumntype{L}[1]{>{\PreserveBackslash\raggedright}p{#1}}

\usepackage{graphicx}
\usepackage{subfigure}
\usepackage{mathrsfs}
\usepackage{hyperref}
\usepackage{url}

\begin{document}

\title{Global-local Enhancement Network for NMFs-aware Sign Language Recognition}

\author{Hezhen Hu}
\affiliation{%
  \institution{University of Science and Technology of China}
  \city{Hefei}
  \postcode{230027}
  \country{China}}
\email{alexhu@mail.ustc.edu.cn}

\author{Wengang Zhou}
\affiliation{%
  \institution{University of Science and Technology of China, Institute of Artificial Intelligence, Hefei Comprehensive National Science Center}
  \country{China}}
\email{zhwg@ustc.edu.cn}

\author{Junfu Pu}
\affiliation{%
  \institution{University of Science and Technology of China}
  \country{China}}
\email{pjh@mail.ustc.edu.cn}

\author{Houqiang Li}
\affiliation{%
  \institution{University of Science and Technology of China, Institute of Artificial Intelligence, Hefei Comprehensive National Science Center}
  \country{China}}
\email{lihq@ustc.edu.cn}

\renewcommand{\shortauthors}{H. Hu, et al.}

\begin{abstract}
Sign language recognition (SLR) is a challenging problem, involving complex manual features, \emph{i.e.,} hand gestures, and fine-grained non-manual features~(NMFs), \emph{i.e.,} facial expression, mouth shapes, \emph{etc}. 
Although manual features are dominant, non-manual features also play an important role in the expression of a sign word. 
Specifically, many sign words convey different meanings due to non-manual features, even though they share the same hand gestures. 
This ambiguity introduces great challenges in the recognition of sign words. 
To tackle the above issue, we propose a simple yet effective architecture called Global-local Enhancement Network~(GLE-Net), including two mutually promoted streams towards different crucial aspects of SLR.
Of the two streams, one captures the global contextual relationship, while the other stream captures the discriminative fine-grained cues. 
Moreover, due to the lack of datasets explicitly focusing on this kind of features, we introduce the first non-manual-features-aware isolated Chinese sign language dataset~(NMFs-CSL) with a total vocabulary size of 1,067 sign words in daily life.
Extensive experiments on NMFs-CSL and SLR500 datasets demonstrate the effectiveness of our method.
\end{abstract}



\maketitle

\section{Introduction}
Sign language, as a commonly used communication way for the deaf community, is characterized by the unique linguistic property. 
It conveys the semantic meaning by both manual (\emph{i.e.}, hand shape, orientation, movement and position) and non-manual features (\emph{i.e.}, facial expressions, eye gaze, mouth shape, \emph{etc}). 
To help the hearing people easily understand sign language and make the daily life of the deaf community more convenient, sign language recognition~(SLR) has been widely studied, which targets at automatically recognizing sign words performed by the signer from a given video.

Basically, sign language recognition can be grouped into two sub-tasks, isolated SLR and continuous SLR. 
The former~\cite{yin2016iterative, huang2018attention, guo2017online} works at the word level, serving as a fundamental task for continuous SLR and providing a test bed for exploring discriminative representation for SLR.
The latter~\cite{koller2019weakly, cui2019deep, pu2019iterative} directly translates the input video sequence into the text sentences.
In this paper, we focus on the former task.

\begin{figure}
\begin{center}
\includegraphics[width=0.9\linewidth]{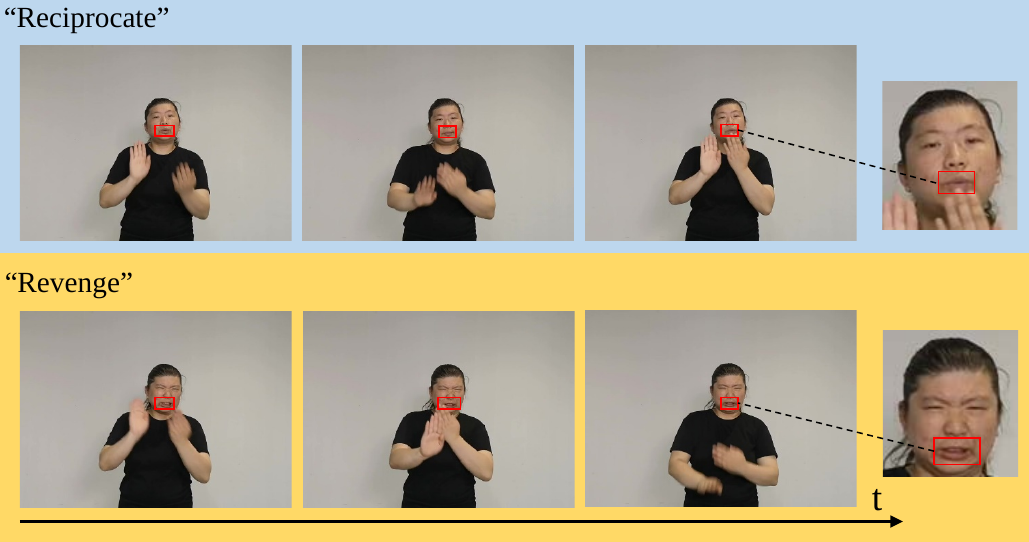}
\end{center}
\caption{An example of the confusing words in CSL. CSL signs ``reciprocate'' (top row) and ``revenge'' (bottom row) have quite diverse semantic meanings. However, they only differ in the facial region with the same hand gesture.}
\label{fig:int}
\end{figure}

Isolated SLR can be viewed as a fine-grained action classification task, where it is significant to extract discriminative feature representations from visual inputs. 
Early approaches tried to tackle this problem with hand-crafted features representing hand gestures, such as HOG~\cite{buehler2009learning, zhang2016chinese} and HGCM~\cite{wang2019novel}.
Recent methods have tried to leverage the advance in the deep learning-based methods for spatio-temporal representations~\cite{huang2015sign, koller2018deep}.
There also exist methods~\cite{camgoz2017subunets, huang2018video, wang2019novel} showing performance gain by using the explicitly cropped hand as an auxiliary assisting cue.
Huang \emph{et al.}~\cite{huang2018attention} present attention-based 3D-Convolutional Neural Networks (3D-CNNs) for SLR.
These methods focus on either the global representation or only the local hand region.

Besides hand motions, non-manual features also play an important role in the expression of a sign word.
Non-manual features include facial expressions, mouth shapes, movements of the head, \emph{etc.}
Similar to the function of the tone in spoken language, non-manual features make the signs much more expressive.
It is worth noting that some sign words share the same gestures but with different non-manual features, especially on the facial region. 
As shown in Figure~\ref{fig:int}, these two words have entirely different meanings, \emph{i.e.,} ``reciprocate'' and ``revenge''. However, they only differ in the motions of the mouth. 
This ambiguity issue, however, is ignored in previous studies on sign language recognition.
Although such ambiguity issue might be resolved by modeling the context of continuous sign language sentence, it is still significant to approach it in a bottom-up paradigm, which will benefit the sentence-level SLR.

To tackle the above issue, we propose a simple yet effective approach called Global-local Enhancement Network~(GLE-Net). 
The key is to enhance the ratio of useful cues to noise for recognition.
Specifically, our approach contains two streams to enhance useful cues for global and local features separately. 
On the one hand, isolated signs consist of complex spatio-temporal contextual relationships among the hand motions and non-manual features, jointly with useless personal attributes, such as clothing. 
Thus we utilize the global enhancement module in a holistic view.
The voxel at every spatio-temporal location will aggregate features from all positions to enhance itself. 
Through this process, this module captures global contextual information and filters out irrelevant personal information.
On the other hand, as mentioned before, some discriminative cues are spatially fine-grained. 
We introduce a local enhancement module by adaptively localizing and emphasizing them. 
Emphasized areas will help the network capture these cues more easily.
Further, we collaboratively learn these two streams via mutual promotion.

Instead of explicitly involving these NMFs-aware confusing words, previous datasets~\cite{chai2014devisign, huang2018attention, ronchetti2016lsa64} focus more on the hand gestures.
For the validation of our method, we include these confusing words and propose a new non-manual-features-aware isolated Chinese sign language dataset~(NMFs-CSL). 
It covers a large vocabulary size of 1,067 sign words.
To our best knowledge, it is the first dataset explicitly emphasizing the importance and challenge of non-manual features in sign language.
It may also serve as a new benchmark for fine-grained video classification.
Extensive experiments on this dataset validate the effectiveness of our method.

The remainder of this paper is organized as follows. 
We first review the related works in Section~\ref{sec:related}.
Then in Section~\ref{sec:dataset}, the details of our proposed NMFs-CSL dataset are described.
After that, we elaborate our approach in Section~\ref{sec:apporach}.
Extensive experimental results are discussed in Section~\ref{sec:experiments}.
Finally, we conclude this paper in Section~\ref{sec:conclusion}.

\section{Related Works}
\label{sec:related}
In this section, we briefly review the related topics, including sign language recognition, attention mechanism and sampling methods. Besides, we summarize the benchmark datasets for isolated sign language recognition. 

\textbf{Sign Language Recognition.}
Sign language recognition~(SLR) can be grouped into two sub-tasks, \emph{i.e.,} isolated SLR and continuous SLR.
Existing approaches for these sub-tasks both show the importance of discriminative feature representations. 
Early works employ hand-crafted features, such as Volume Local Binary Patterns (VLBP)~\cite{wang2014similarity}, HOG or HOG3D based features~\cite{buehler2009learning, cooper2012sign, koller2015continuous, liwicki2009automatic}, SIFT based features~\cite{tharwat2015sift, yasir2015sift} and motion trajectories~\cite{evangelidis2014continuous, koller2015continuous} to represent hand gestures.
Wang \emph{et al.}~\cite{wang2019novel} propose a hierarchical Grassmann covariance matrix (HGCM) model for sign description.
Hidden Markov Models~(HMM) is also utilized for modeling the temporal relationship~\cite{starner1995visual, zhang2016chinese}.

Recent advances in computer vision tasks show the high capacity of CNN for discriminative feature representations. 
In the related task of action recognition, many works have explored the design of deep architectures based on 2D-CNN, 3D-CNN and the mixture of them~\cite{carreira2017quo, chen2018multi, nie2017enhancing, Qiu_2017_ICCV, qiu2019learning, simonyan2014two, wang2016temporal, xie2018rethinking, zhou2018mict}.
TSN~\cite{wang2016temporal} proposes a novel framework using 2D-CNN as the backbone. 
It is based on the idea of long-range temporal structure modeling. 
As one of the most popular 3D-CNNs, Inflated 3D-CNN~(I3D)~\cite{carreira2017quo} inflates all the 2D kernels in the original GoogLeNet~\cite{szegedy2015going} into the 3D one and achieves a large performance gain over previous methods.
Slowfast~\cite{feichtenhofer2019slowfast} is a state-of-the-art method in general action recognition, which capitalizes two paths at different frame rates, \emph{i.e.,} fast path for temporal
information and slow path for spatial semantics.
General video classification methods try to distinguish the complex actions with a large diversity of background, temporal duration, \emph{etc.}
In contrast, sign language recognition focuses on the manual and non-manual features of the signer.

For the task of sign language recognition, various CNN models have been studied~\cite{cui2019deep, koller2019weakly, koller2017re, liu2017continuous, pu2018dilated,  pu2019iterative, shi2019fingerspelling, zhou2019dynamic}.  
Koller~\emph{et al.} utilize 2D-CNN with HMMs~\cite{koller2016deep, koller2018deep}. 
Huang~\emph{et al.}~\cite{huang2018attention} embed 3D-CNNs to learn spatio-temporal features from the raw videos with a spatial and temporal attention mechanism. 
These methods mainly focus on the global structure or the local hand region, which is insufficient to capture fine-grained non-manual features.
In this paper, we focus on the isolated SLR and take the non-manual features into consideration.

\textbf{Attention mechanism and Sampling methods.}
Attention mechanism enables the model to focus on the discriminative cues and filter out irrelevant disturbance. 
It has been successfully applied to many computer vision tasks, such as image classification, image captioning, video classification, \emph{etc.}
Hard attention makes a hard selection on the region containing useful cues~\cite{mnih2014recurrent, gkioxari2015contextual, mallya2016learning, fu2017look}.
R*CNN~\cite{gkioxari2015contextual} uses auxiliary region to encode context with the human.
Mnih~\emph{et al.}~\cite{mnih2014recurrent} propose to extract information by selecting a sequence of regions based on reinforcement learning.
Soft attention replaces the hard selection with weighted sum~\cite{xu2015show, li2018videolstm, sharma2015action}.
Li~\emph{et al.}~\cite{li2018videolstm} propose an end-to-end sequence learning architecture based on the soft-attention LSTM~\cite{sharma2015action}, which explores the spatial correlations and motion to guide the attention.
This kind of attention usually needs extra information as guidance and may not deal with the long-range dependencies well.
Self-attention, firstly proposed in machine translation~\cite{vaswani2017attention}, tries to tackle this issue by computing and aggregating weighted features from all positions. 
Recently, several works have applied it in computer vision tasks~\cite{hu2018relation, zhang2018self, wang2018non}. Non-local network~\cite{wang2018non} is the first to utilize it in the image and video tasks, and shows notable performance gain by capturing the global contextual relationship.

Non-uniform sampling methods have been applied in fine-grained image tasks as a pre-processing method to the raw images~\cite{jaderberg2015spatial, recasens2018learning,Zheng2019looking}. 
It adaptively enlarges certain areas of the original input image, and helps CNN capture discriminative fine-grained details more easily, thus bringing notable performance gain.
SSN~\cite{recasens2018learning} first uses the saliency map as guidance in the sampling process. 
To reduce the spatial distortion, TASN~\cite{Zheng2019looking} proposes to decompose the saliency map into the x and y axes.
Different from the previous methods, we propose to use the sampling module on feature maps for emphasizing the discriminative regions. 
Further, we notice that the essence of the above mentioned two mechanisms is enhancing the ratio of useful cues to noise from different aspects. 
This also motivates us to take a follow-up exploration towards the incorporation of these mechanisms and formulate our enhancement network.

\textbf{Isolated Sign Language Datasets.} 
For decades, datasets of isolated sign languages have been collected with different sensors, such as data gloves, Leap Motion, Kinect and monocular RGB cameras, \emph{etc.} 
The languages covered by different datasets include American, Chinese, German, Polish, \emph{etc.}
We summarize these datasets in Table~\ref{overview}.

In early ages, with glove based devices, most datasets directly capture the detailed motion status of each finger articulation and the whole hand~\cite{fels1993glove, wang2001real, stefanidis20203d, avola2018exploiting}.
Kim \emph{et al.} \cite{kim1996dynamic} propose a dataset covering 25 words in Korean sign language.
It is captured by the VPL data glove, which records hand positions, orientations and movements.
The emergence of the non-contact motion sensing devices, \emph{e.g.} Leap Motion and RealSense, enables it to capture hand motions without wearable gloves.
Leap Motion Controller~(LMC) is able to track and model hands at 200 frames per second~(fps).
It records a total of 28 kinds of features, containing 3D joint locations and orientations.
Chuan \emph{et al.}~\cite{chuan2014american} record a dataset with 26 letters in the American alphabet by LMC.
However, non-contact motion sensing devices only capture accurate hand motions, but ignore other crucial features, \emph{e.g.} body pose, non-manual features, \emph{etc.}
Hence, researchers choose the vision based sensors for data collection, including Kinect, ToF, monocular RGB cameras, \emph{etc.}
These datasets usually cover a larger number of vocabulary and videos compared with the former.
Kinect records multiple modalities, including RGB, skeleton and depth data.
SLR500~\cite{huang2018attention} is one of the datasets recorded by Kinect, covering 125,000 videos with the vocabulary size of 500.
Recently, there also exist datasets recorded only on the monocular RGB cameras, such as MS-ASL~\cite{joze2018ms}, which facilitate SLR into the real-world application.

In summary, these datasets have their own properties and focus more on the hand gestures.
As mentioned above, it also should be notable that there exists ambiguity between words introduced by the non-manual features.
However, few datasets explicitly focus on them.
To tackle this issue, we propose a new dataset explicitly emphasizing the importance of the non-manual features.

\begin{table*}
\centering
\small
\tabcolsep=10pt
\begin{tabular}{l|ccccc}
\hline
Datasets & \#Signs & \#Videos  & Device & Language & Year  \\ \hline \hline
KSL~\cite{kim1996dynamic} & 25 & - & Data Glove & Korean & 1996 \\
HMU~\cite{holden2001visual} & 22 & 44 & Color Glove & Australian & 2001 \\
MSL-Alphabet~\cite{chuan2014american} & 26 & 104 & Leap Motion & American & 2014 \\
CSL-Alphabet~\cite{huang2015sign2} & 26 & 78 & RealSense & Chinese & 2015 \\
RWTH-BOSTON-50~\cite{zahedi2005combination} & 50 & 483 & Multi-view Cam & American & 2005 \\
DGS-Kinect-40~\cite{cooper2012sign} & 40 & 3,000 & Kinect & German & 2012 \\
PSL ToF~\cite{kapuscinski2015recognition}  & 84 & 1,680  & ToF  & Polish  & 2013  \\
PSL Kinect~\cite{oszust2013polish} & 30 &  300   & Kinect & Polish & 2013 \\
DEVISIGN-L~\cite{chai2014devisign} & 2,000 & 24,000 & Kinect & Chinese & 2014 \\
LSA64~\cite{ronchetti2016lsa64}    & 64      & 3,200    & Monocular RGB & Argentinian    & 2016  \\
SLR500~\cite{huang2018attention}   & 500 & 125,000 & Kinect & Chinese & 2018 \\
MS-ASL~\cite{joze2018ms}   & 1,000   & 25,513   & Monocular RGB     & American      & 2019  \\
Ours     & 1,067   & 32,010   & Monocular RGB    & Chinese       & 2020 \\ \hline
\end{tabular}
\caption{An overview of the isolated sign language datasets in different languages.}
\label{overview}
\end{table*}

\begin{figure*}
\begin{center}
\includegraphics[width=1\linewidth]{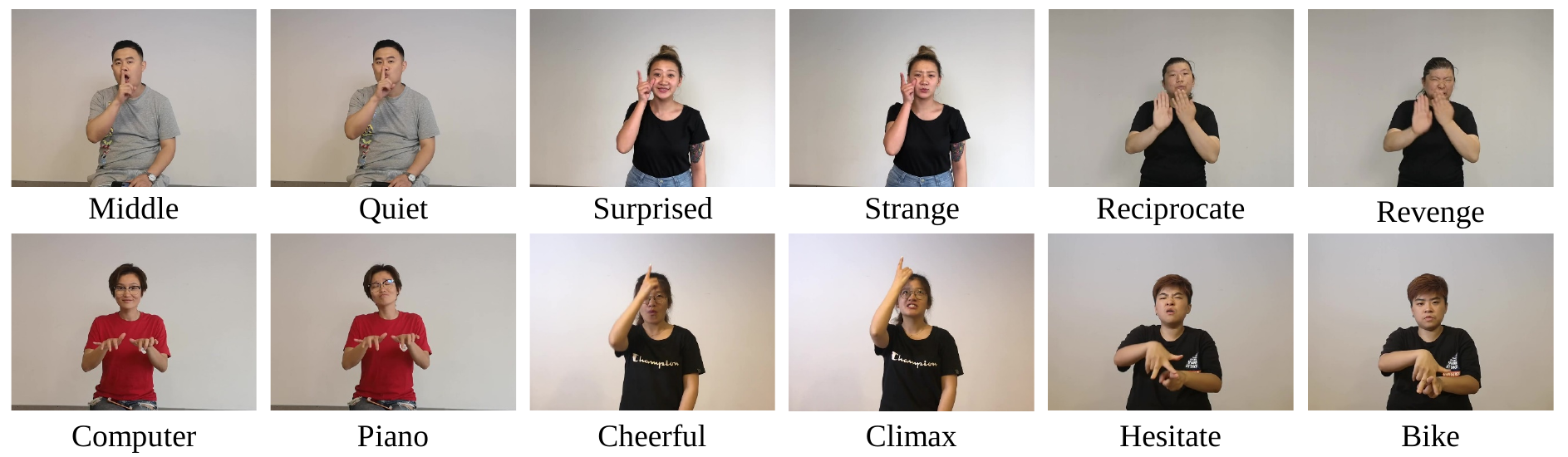}
\end{center}
\caption{Some examples from the NMFs-CSL dataset. Each video is represented by one frame and the confusing words are listed coupled. Better viewed by zooming in.}
\label{fig:dataset}
\end{figure*}

\section{NMFs-CSL Dataset}
\label{sec:dataset}
In this section, we introduce our proposed non-manual-features-aware isolated Chinese sign language dataset~(NMFs-CSL).\footnote{\url{http://home.ustc.edu.cn/~alexhu/Sources/index.html}}
We first describe the major characteristics of our datasets. 
Then we detail the construction procedures and statistical analysis of our dataset.
Finally, we summarize the challenges in this dataset.

As stated before, previous datasets focus more on the hand gestures. 
Although hand gestures are more dominant, the non-manual feature is also an essential part in expressing sign language. 
Non-manual features include mouth morphemes, eye gazes, facial expressions, body shifting, and head tilting.
As for the word-level, it is notable that there are a lot of daily words with different meanings, sharing the same gestures but having different non-manual features, especially on the facial region. 
Accurate recognition of these words is meaningful to the disambiguation and linguistic research. 
Thus we collect these words in our dataset. 
Some samples of them are shown in Figure~\ref{fig:dataset}.
It can be observed that the NMFs are able to change the word to its totally reverse meaning, such as `reciprocate' and `revenge'.
Besides, NMFs can also indicate the intensity of the emotion, such as `surprised' and `strange'. 
To the best of our knowledge, we are the first to explicitly consider the role of non-manual features by including these confusing words.

\begin{figure}[H]
\centering
\subfigure[Histogram of frame numbers]{
\label{a}
\begin{minipage}[b]{0.49\linewidth}
\centering
\includegraphics[width=0.75\linewidth]{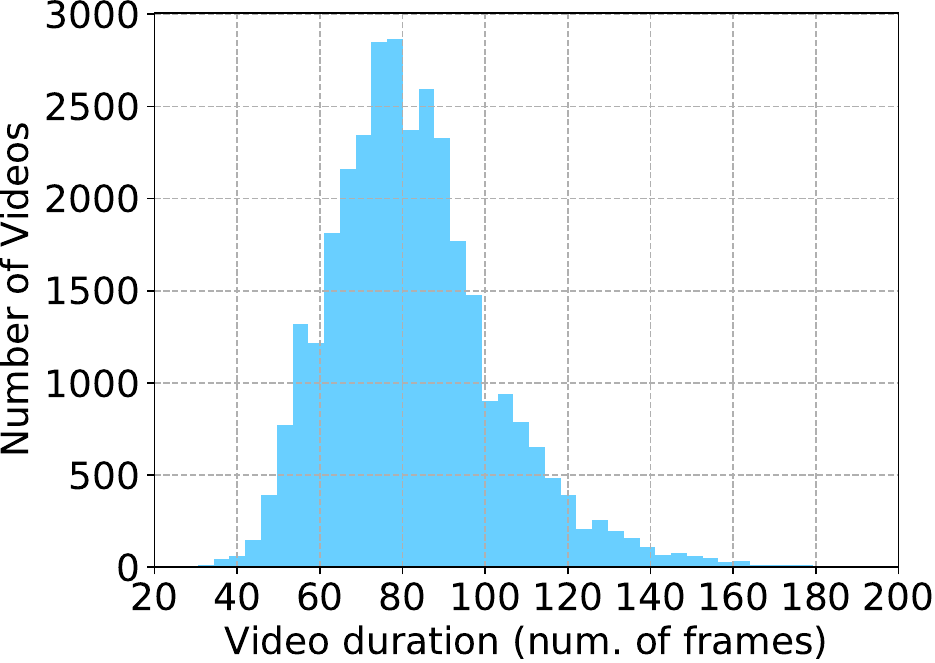}\vspace{1pt}\end{minipage}}
\subfigure[Pie chart on word classes]{
\label{b}
\begin{minipage}[b]{0.47\linewidth}
\centering
\includegraphics[width=0.75\linewidth]{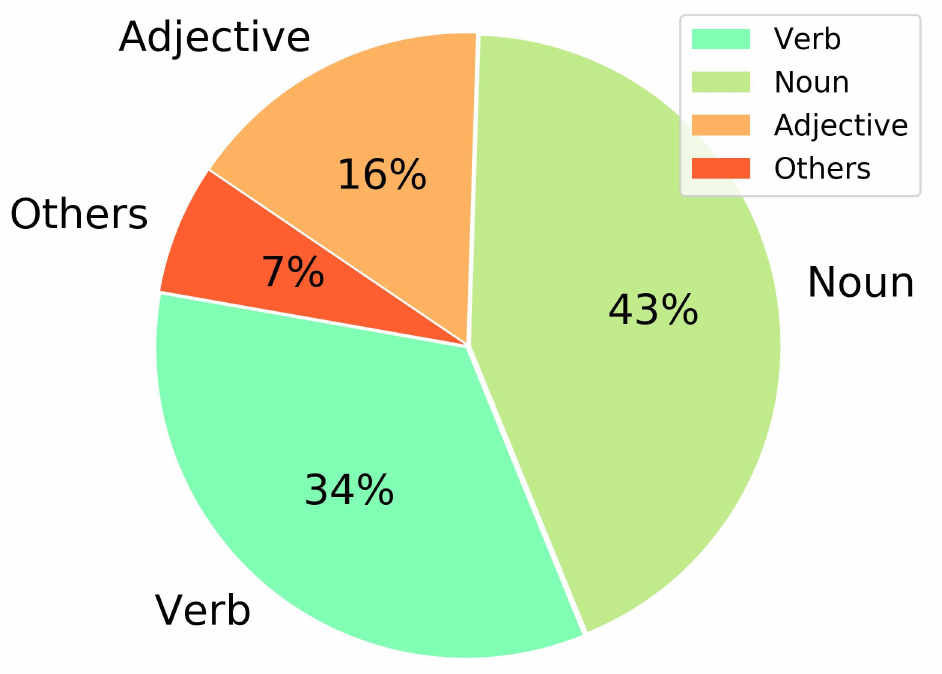}\vspace{1pt}\end{minipage}}
\caption{Detailed statistical analysis of NMFs-CSL dataset.}
\label{stat}
\end{figure}

Since we aim to leverage the minimal hardware requirement for sign language recognition, our dataset only consists of monocular RGB videos. 
In this way, recognition approaches do not rely on special equipment, and it will facilitate the design of tools for communication with the deaf community. 
Our dataset is recorded in a background controlled environment by the camera on the portable mobile phone.
The videos are recorded at 30 frames per second~(fps) with the shorter spatial size as 720 pixels. 
Besides, the signers' wear and background lighting conditions are varying. 

The dataset contains 32,010 samples in total with a vocabulary size of 1,067.
In this dataset, there are 610 NMF-aware confusing words.
To make the scenario more realistic, each sign word is performed 3 times by 10 signers in near-frontal views, thus containing 30 samples.
The total duration of all samples is 24.4 hours.
The comparison of our dataset with its counterparts is shown in Table~\ref{overview}. 
The detailed statistical analysis is shown in Figure~\ref{stat}. 
It can be observed that frame numbers of video samples usually range from 50 to 120.
Words contained in this dataset largely distribute in verb and noun. 

The challenges of NMFs-CSL dataset are summarized in three aspects. 
First, it is a large-scale isolated sign language dataset with a vocabulary size of over 1,000. 
Second, we explicitly include 610 NMFs-aware confusing words. With these words, the inter-class similarity increases notably, and more fine-grained discriminative cues exist, which are crucial for the accurate recognition.
Compared with current fine-grained datasets, its total number of classes is larger.
Different from them, \emph{e.g.,} Something-Something~\cite{goyal2017something} focusing on human-object interaction, fine-grained cues in NMFs-CSL dataset are humam-centric.
Third, when performing the same sign, especially on the expression of non-manual features, there exist inter-signer variations.

\begin{figure*}
\begin{center}
\includegraphics[width=1\linewidth]{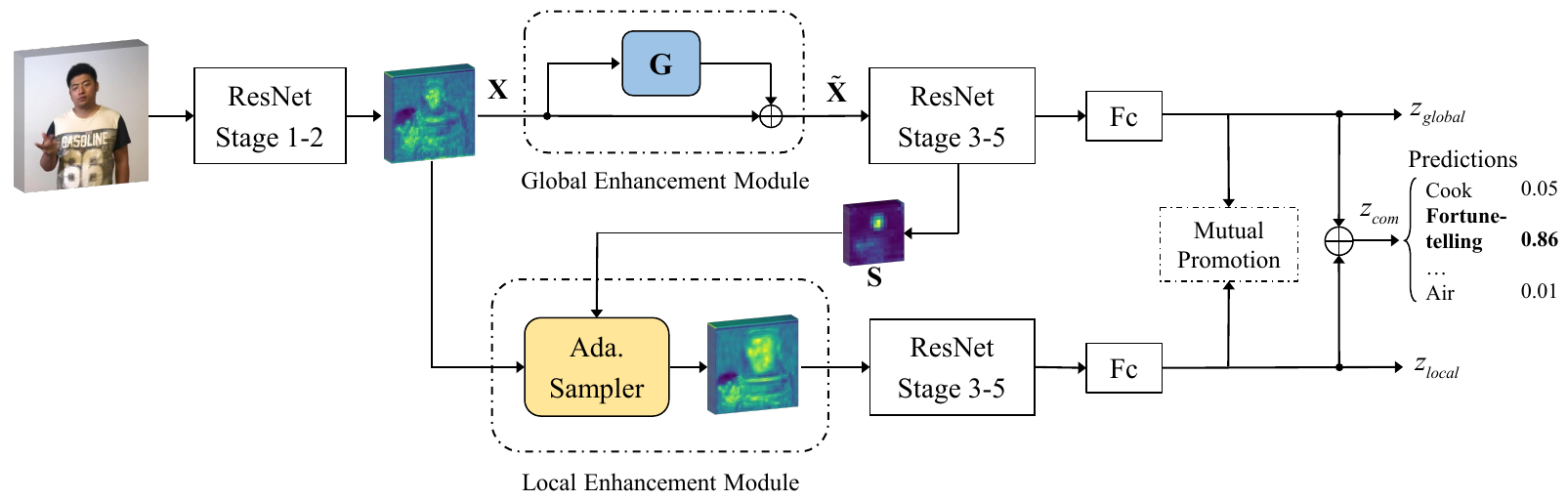}
\end{center}
\caption{Illustration of the Global-local Enhancement Network. The input video first goes through several convolution layers for feature representations. Then there are two paths for capturing discriminative cues, \emph{i.e.,} the \emph{global} stream (upper one) for enhancing the global contextual relationship, the \emph{local} stream (lower one) for emphasizing fine-grained details. These two paths work in mutual promotion to enhance each other. Finally, the prediction is performed by sum fusion of these two branches.}
\label{fig:overall}
\end{figure*}

\section{Our Approach}
\label{sec:apporach}
In this section, we first present our general framework, which consists of two cooperative enhancement modules. 
Then we introduce these two enhancement modules for global and local information, respectively. 
Finally, we discuss how to promote them collaboratively and aggregate them for the final prediction.

\subsection{Framework Overview}
Signs cover complex spatio-temporal contextual relationships among the manual and non-manual features.
Jointly with this global relationship, there also exist discriminative cues hidden in the fine-grained details.
This nature introduces great challenges in isolated SLR.
To tackle these issues, the key is to enhance useful cues over noise for global and local features.
With the above motivation, we propose a Global-local Enhancement Network~(GLE-Net).

The overall framework is illustrated in Figure~\ref{fig:overall}. 
Given a video of isolated sign language, it first goes through several convolutional layers of the backbone, \emph{i.e.,} ResNet stage 1 and 2 in our implementation, to extract feature representations. 
Then it branches into two streams, \emph{i.e.,} the \emph{global} stream for context information and the \emph{local} stream for fine-grained cues. 
Each stream has its own customized enhancement module for capturing complementary discriminative feature representations. 
The outputs of the two streams $z_{global}$ and $z_{local}$ are supervised by the standard cross-entropy loss, denoted as $\mathcal{L}_{global}$ and $\mathcal{L}_{local}$, respectively. 
Then to further benefit the learning of each stream, a mutual promotion loss $\mathcal{L}_{mu}$ is utilized.
Furthermore, the results from two streams are fused by summation to make the final prediction $z_{com}$, which is also supervised by the cross-entropy loss $\mathcal{L}_{com}$.
Finally, we combine the above-mentioned losses to formulate the final training loss as follows,
\begin{equation}
    \mathcal{L} = \lambda(\mathcal{L}_{com} + \mathcal{L}_{global} + \mathcal{L}_{local}) + (1 - \lambda) \mathcal{L}_{mu}.
\label{full_obj}
\end{equation}
Since $\mathcal{L}_{com}$, $\mathcal{L}_{global}$ and $\mathcal{L}_{local}$ are all implemented as the cross-entropy loss, we equally combine them and perform weighted sum with $\mathcal{L}_{mu}$.

\subsection{Global Enhancement Module}
Sign videos contain complex motion cues with the contextual relationship.
These cues may not be well captured by conventional convolution, which operates on a window of neighboring pixels.
Besides, there also exists disturbance to these cues, such as irrelevant personal information.
To tackle the limitation of convolution and filter out disturbance, we aim to enhance these contextual cues in a holistic view and formulate our global enhancement module.
Specifically, it contains the enhancement feature extractor $\mathbf{G}$ with the residual connection. 

Motivated by the previous work~\cite{wang2018non}, we utilize the self-attention module in our global enhancement module.
Given a feature map $\mathbf{X}\in{{\mathbb{R}}^{C\times T\times H\times W}}$, where $C, T, H$ and $W$ denote the channel, temporal and spatial (height and width) dimension of the feature map, respectively.
We first feed it into independent convolutional layers to generate three new feature maps with the same shape as $\mathbf{X}$ and reshape them into ${{\mathbb{R}}^{C\times N}}$, where $N=THW$ to get $\mathbf{A}$, $\mathbf{B}$ and $\mathbf{C}$, respectively. 
After that, we use matrix multiplication to calculate the semantic similarity between voxels and apply a softmax layer for normalization, which defines the enhancement map $\mathbf{E} \in {{\mathbb{R}}^{N\times N}}$,
\begin{equation}
    {{\mathbf{E}}_{ij}}=\frac{\exp ({{\mathbf{A}}_{i}}\cdot{{\mathbf{B}}_{j}})}{\sum\limits_{j=1}^{N}{\exp ({{\mathbf{A}}_{i}}\cdot{{\mathbf{B}}_{j}})}},
\end{equation}
where $\mathbf{A}_i$ and $\mathbf{B}_j$ denote the feature at the $i^{th}$ and $j^{th}$ location of the feature map, respectively. 
$\mathbf{E}_{ij}$ denotes the impact from the $j^{th}$ location to the $i^{th}$ location. 

With the enhancement map, we embed the capacity to enhance every pixel using long-range global dependencies. 
After that, we perform matrix multiplication between $\mathbf{C}$ and $\mathbf{E}$, and reshape the result to get a feature map ${{\mathbb{R}}^{C\times T\times H\times W}}$. 
In this way, we aggregate the features from $\mathbf{C}$ using the weighting matrix $\mathbf{E}$. 
Finally, we get the enhanced feature map $\tilde{\mathbf{X}}\in{{\mathbb{R}}^{C\times T\times H\times W}}$ by the following formulation,
\begin{equation}
    \tilde{\mathbf{X}}=\beta f(\mathbf{C}\mathbf{E})+\mathbf{X},
\end{equation}
where $f(\cdot)$ denotes the reshaping operation and we introduce the residual connection and initialize it as zero by setting $\beta=0$~\cite{he2016deep}. 
Then $\beta$ is updated via back-propagation during training.
By aggregating similar semantic features, the feature at each spatio-temporal location is enhanced. 
Discriminative contextual cues are enhanced mutually while irrelevant information will be compressed.

\subsection{Local Enhancement Module}
Besides complex motion cues with contextual relationship, there also exist discriminative fine-grained cues in sign videos.
These fine-grained cues usually include the mouth shape, eye gaze, facial expressions or a mixture of them.
They cover small spatial regions in the spatio-temporal tubes and vary along the time.
Thus they easily disappear in repeated convolution and pooling operations.
In order to preserve such fine-grained details, we introduce a local enhancement module through adaptive sampling to highlight them with higher resolution.

\begin{figure}[t]
\begin{center}
\includegraphics[width=0.5\linewidth]{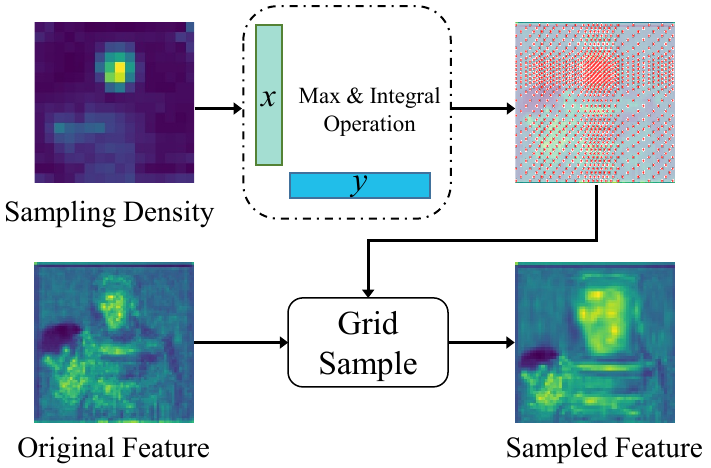}
\end{center}
\vspace{-0.4cm}
\caption{Illustration of the adaptive sampling in the local enhancement module.}
\label{fig:local}
\end{figure}

Given a feature map $\mathbf{X}\in{{\mathbb{R}}^{C\times T\times H\times W}}$, we intend to perform adaptive sampling spatially as shown in Figure~\ref{fig:local}. 
We take one feature map at the specific time $t$ as an example, denoted as $\mathbf{X}_{t}$. 
The adaptive sampling takes the saliency map $\mathbf{S}\in{{\mathbb{R}}^{H\times W}}$ as guidance, where the area with larger saliency value will be sampled more densely~\cite{Zheng2019looking}. 
Specifically, we first calculate the integral of the saliency map over $x$ and $y$ axes and do normalization as follows,

\begin{equation}
\begin{split}
    {{P}_{x}}(k_x)={\sum\limits_{i=1}^{k_x}{\max ({{\mathbf{S}}_{i*}})}/\sum\limits_{i=1}^{W}{\max ({{\mathbf{S}}_{i*}}})}, \\
    {{P}_{y}}(k_y)={\sum\limits_{j=1}^{k_y}{\max ({{\mathbf{S}}_{*j}})}/\sum\limits_{j=1}^{H}{\max ({{\mathbf{S}}_{*j}})}},
\end{split}
\end{equation}
where $k_x\in[1, W]$ and $k_y\in[1,H]$. 
In this way, we get the distribution functions over the $x$ and $y$ axes. 
Then we accomplish the sampling by the inverse function of the distribution function as follows,
\begin{equation}
    {{\mathbf{O}}_{ij}}=({{\mathbf{X}}_{t}}){_{P_{x}^{-1}(i),P_{y}^{-1}(j)}},
\end{equation}
where $\mathbf{O}$ is the sampled feature map at time $t$.

The above operation involves a saliency map $\mathbf{S}$, which is generated using the high-level feature in the global stream as reference. 
Specifically, we choose the commonly used trilinear attention as follows,
\begin{equation}
\label{saliency}
    \mathbf{S}=mean(f(f(\mathbf{Y}_t){{\mathbf{Y}_t}^{T}})\mathbf{Y}_t),
\end{equation}
where $\mathbf{Y}_t\in{{\mathbb{R}}^{{C_2}\times {N_2}}}$ denotes the corresponding reshaped high-level feature in the global stream at the same time as $\mathbf{X}_t$, $f(\cdot)$ denotes the softmax normalization function, and $mean(\cdot)$ is the average function through the channel dimension.
The trilinear attention also aims to capture contextual relationship and the average function is intended to produce a robust saliency map.
The output $\mathbf{S}\in{{\mathbb{R}}^{1\times {N_2}}}$ is finally reshaped and upsampled to the same size as original feature map $\mathbf{X}_t$.

By resampling the feature map, the fine-grained cues are localized and emphasized.
Thus they will be easily captured by the convolution operation.

\subsection{Mutual Promotion Module}
The two modules mentioned above enhance the feature representations for SLR from two complementary aspects.
One aims to capture the long-range contextual dependency, while the other targets at emphasizing the discriminative fine-grained details.
However, the latter inevitably brings information loss on the global context and fails to give feedback to the former.

To tackle the above issue, we propose the mutual promotion module, which optimizes two enhancement modules in a collaborative way. 
Inspired by the mutual learning~\cite{zhang2018deep}, we intend to make each module learn not only from the current label but also the predicted probability distribution from its peer. 
Given the output prediction after the softmax layer from two streams, denoted as $p_{1}$ and $p_{2}$, we measure the match of the two streams' predictions by adopting Kullback Leibler~(KL) divergences as follows,
\begin{equation}
\begin{split}
    {{D}_{KL}}({{p}_{2}}\parallel {{p}_{1}})=\sum\limits_{m=1}^{M}{p_{2}^{m}\log \frac{p_{2}^{m}}{p_{1}^{m}}}, \\
    {{D}_{KL}}({{p}_{1}}\parallel {{p}_{2}})=\sum\limits_{m=1}^{M}{p_{1}^{m}\log \frac{p_{1}^{m}}{p_{2}^{m}}}, \\
\end{split}
\end{equation}
where $M$ is the total number of classes.
We denote a summation of these two KL divergences as the mutual promotion loss $\mathcal{L}_{mu}$ as follows, 
\begin{equation}
    \mathcal{L}_{mu} = {{D}_{KL}}({{p}_{2}}\parallel{{p}_{1}}) + {{D}_{KL}}({{p}_{1}}\parallel{{p}_{2}}).
\end{equation}

Although two streams enhance the feature representations from two aspects and are implemented differently, they both aim to classify the same video correctly.
With the mutual promotion loss added, it aligns each stream's class posterior with the class probability of the other stream.
In this way, each stream refer to the another stream during training and converges to a more robust minimal.
By referring to each other, the connections between two streams are also established, with more information shared between each other.
The local stream can also implicitly adjust the sampling process.
Meanwhile, the modified emphasized fine-grained areas will help perform better classification jointly with the global stream.

\section{Experiments}
\label{sec:experiments}
In this section, we first introduce our benchmark datasets, evaluation metrics and implementation details. 
Then we perform a series of ablation studies on the effects of two enhancement modules and mutual promotion module. 
Finally, we report and analyze the results.

\subsection{Experiment Setup}
In this work, we perform experiments on two benchmark datasets, \emph{i.e.,} NMFs-CSL and SLR500~\cite{huang2018attention}.
For NMFs-CSL, we adopt the signer-independent setting.
8 signers are selected for training with 2 signers for testing. 
For SLR500, as stated before, it is recorded by Kinect, providing RGB, depth and skeleton modalities.
This dataset contains a total vocabulary size of 500 words in daily life.
Different from ours, it does not explicitly include these NMFs-aware confusing words in the dataset. 
The signer-independent setting is also adopted, with 36 signers for training and 14 signers for testing.
Notably, for NMFs-CSL and SLR500, the vocabulary in the training set is equal to that in the testing set.

We evaluate the model using the accuracy metrics. 
More precisely, we compute the top-$K$ accuracy where $K = \left\{1, 2, 5\right\}$ for NMFs-CSL, and $K=1$ for SLR500 dataset. 
In the top-$K$ accuracy, if any of the top-$K$ probability predictions of the sample is consistent with the label, the sample is recognized true.

Our models are all implemented on PyTorch framework and trained on GTX 1080-Ti. 
We choose 2D-CNN and 3D-CNN as our baseline methods, which both use ResNet-50~\cite{he2016deep} as the backbone for a fair comparison. 
All backbones are pretrained on ImageNet~\cite{deng2009imagenet} or Kinetics~\cite{kay2017kinetics} dataset. 
Our 3D-CNN example architecture~\cite{Qiu_2017_ICCV} first inflates 2D $7\times7$ conv1 kernel to $5\times7\times7$ by common practice introduced in I3D~\cite{carreira2017quo}. 
Then each $3\times3$ kernel is replaced with one $1\times3\times3$ spatial convolution and one $3\times1\times1$ temporal convolution. 
In our proposed network, to make trade-off on memory and computational cost, we set the first two stages of ResNet backbone as the sharing feature extractor for the global and local stream.
The source code will be available at \url{https://github.com/AlexHu123/GLE-Net}.

\textbf{Optimization parameters.} 
All the models are trained with the Stochastic Gradient Descent (SGD) optimizer. 
We set weight decay to 1e-4, momentum to 0.9, and dropout to 0.5.
The learning rate is initialized as 5e-3, and reduced by a factor of 0.1 when validation loss saturates. 
Our batch size is as large as the hardware is able to permit.

\textbf{Input size and data augmentation.} 
We first resize all videos to their shorter spatial size as 256 pixels.
During training, we augment samples with random horizontal flipping, random cropping of $224\times224$ pixels spatially. 
During testing, we augment the samples with center cropping of $256\times256$ pixels spatially. 
To reduce the temporal redundancy at the current frame rate, we extract one frame every one interval. 
For training and testing of 3D-CNN, we extract 32 frames temporally by random and center selecting, respectively. 
For its counterpart 2D-CNN, we follow the similar setting introduced in TSN~\cite{wang2016temporal}. 
A sample is divided into 16 segments. 
We constitute totally 16 frames by randomly selecting one frame or selecting the first frame in each segment during training and testing, respectively.

\textbf{Training and testing settings.} During training, we follow the two-stage optimization. 
First, we train the global stream and local stream separately. Then we add the mutual promotion module and jointly finetune the whole framework trained at the first stage. 
During inference, we report the results using the single-crop setting. 

\subsection{Ablation Study}
We perform ablation studies on the hyper parameter and the effectiveness of loss terms in our framework.

\textbf{Hyper parameter $\lambda$.}
\begin{figure}
\begin{center}
\includegraphics[width=0.6\linewidth]{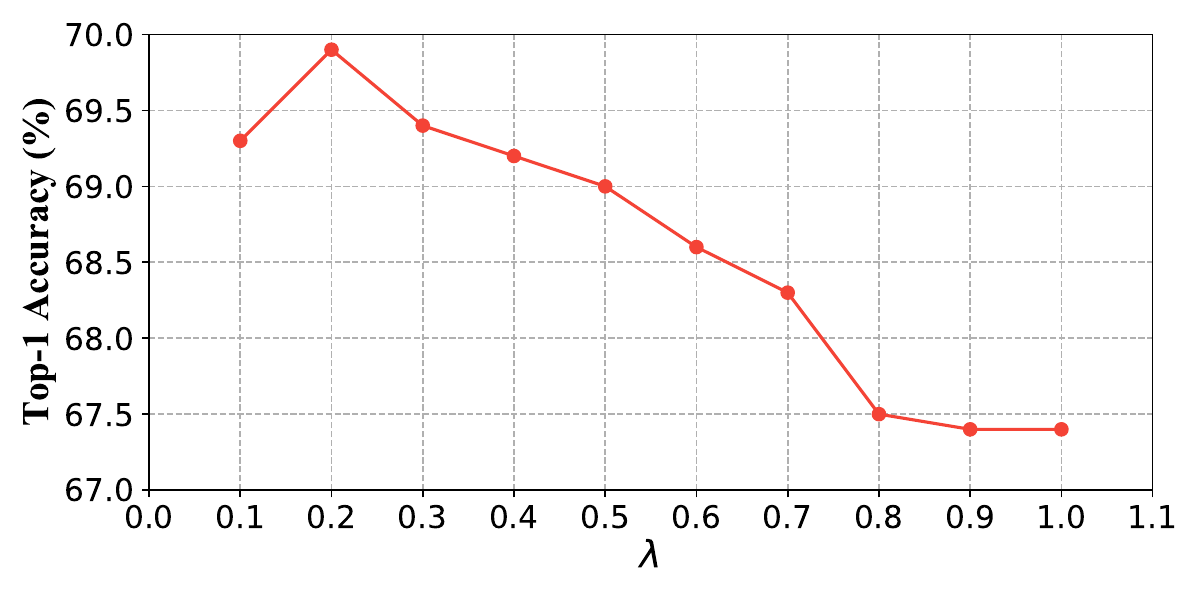}
\end{center}
\caption{Effect of the hyper parameter $\lambda$ in Equation~\ref{full_obj} on NMFs-CSL.}
\label{fig:lambda}
\end{figure}
We study the effect of $\lambda$ in Equation~\ref{full_obj} as shown in Figure~\ref{fig:lambda}.
The experiments are conducted on NMFs-CSL and we utilize the top-1 accuracy as the performance indicator.
The top-1 accuracy reaches the peak value when $\lambda = 0.2$. 
When $\lambda = 0$, the whole model does not converge and we do not include it in the figure.
Unless stated, we utilize $\lambda = 0.2$ as the default hyper parameter.

We implement our global and local enhancement module in their corresponding streams jointly with the mutual promotion module for further performance boosting. 
To verify the effectiveness of each loss terms supervising the framework, we conduct experiments with different settings.

\begin{table}
\small
\begin{center}
\tabcolsep=8.5pt
\begin{tabular}{lccccc}
\hline
Location  & -       & pool2         & res3  & res4  & res5 \\ \hline \hline
Top-1     & 62.1    & \textbf{64.8} & 63.2  & 63.4  & 63.5 \\
Top-2     & 73.2    & \textbf{76.1} & 74.0  & 74.7  & 74.8 \\
Top-5     & 82.9    & 84.7          & 84.2  & 84.5  & \textbf{84.8} \\
\hline
\end{tabular}
\end{center}
\caption{Results of the global enhancement module placed on different locations. The second column indicates our 3D-ResNet-50 backbone as the baseline. (Top-1, Top-2 and Top-5 accuracy (\%)).}
\label{globalen}
\end{table}

\textbf{Global enhancement module.} 
We perform an ablation study on the effects of the global enhancement module placed on different locations. 
We only keep the global enhancement module in the framework and insert it right after the last residual block of a stage.
Considering the high computation cost, we do not add the module at the first or second stages.
As shown in Table~\ref{globalen}, the improvement brought by the module on pool2 is the largest, which leads to 2.7\% top-1 accuracy improvement over the baseline. 
However, the performance gains on res3, res4 and res5 are similar and all smaller than that on pool2. 
One possible explanation is that discriminative cues may be lost by repeated convolution and pooling operations, so the module needs to be added in the front stage to enhance the global contextual information.
We fix the location of the global enhancement module right after the pool2 unless stated in the following experiments.

\begin{table}
\small
\begin{center}
\tabcolsep=12pt
\begin{tabular}{lccccc}
\hline
Location    & -     & res3  & res4           & res5  \\ \hline \hline
Top-1       & 62.1  & 63.2  & \textbf{63.9}  & 63.3   \\
Top-2       & 73.2  & 72.7  & \textbf{73.8}  & 73.1   \\
Top-5       & 82.9  & 81.6  & \textbf{83.0}  & 82.4   \\ \hline
\end{tabular} 
\end{center}
\caption{Effects of selecting different stages in the global stream as reference for saliency map generation. We quantify the effects by reporting the accuracy of the local stream. The second column indicates our 3D-ResNet-50 baseline. (Top-1, Top-2 and Top-5 accuracy (\%)).}
\label{localen}
\end{table}

\begin{figure*}
\begin{center}
\includegraphics[width=1\linewidth]{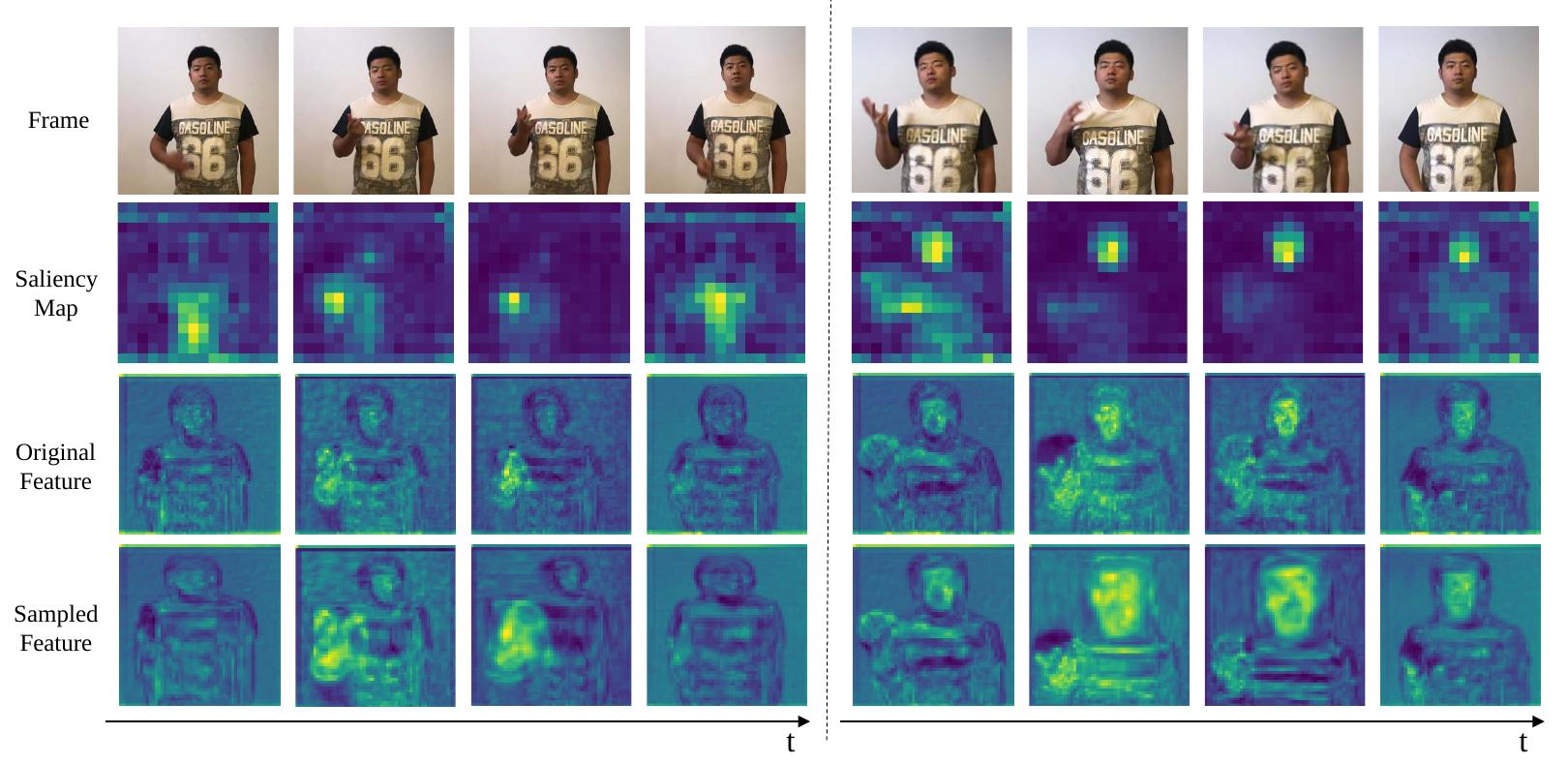}
\end{center}
\vspace{-0.3cm}
\caption{Visualization of the adaptive sampling on the feature map. The two parts correspond to two confusing words, \emph{i.e.}, ``walnut'' (left) and ``fortune-telling'' (right). They share the same hand gesture but only differ in the head movement. We plot each sample along the time. The corresponding video frame, saliency map~(according to Equation~\ref{saliency}), original feature map and sampled feature map are shown in the order of row.}
\label{fig:vis}
\end{figure*}

\textbf{Local enhancement module.} 
An ablation study is then performed on the local enhancement module to investigate its specific setting.
From Table~\ref{localen}, we can observe that using the feature map right after the res4 stage as reference to generate the saliency map achieves the largest improvement over the baseline, with a 1.8\% top-1 accuracy gain. 
In contrast, its counterparts, res3 and res5 bring smaller accuracy improvement. It can be explained that using the high-level feature as the reference can help generate a more discriminative saliency map to guide the sampling process. 
However, when it goes to the top layers of the networks, the feature map is more semantically abstract with a smaller spatial size. 
A too small spatial size leads to the resolution degradation, which hinders the accurate sampling for discriminative cues and finally weakens the performance improvement. 
As a result, we choose the feature map right after the res4 in the global stream as reference to generate the saliency map.

Besides, visualization on the effects of the adaptive sampling process is shown in Figure~\ref{fig:vis}. 
It shows the example of two confusing words, \emph{i.e.}, ``walnut'' and ``fortune-telling''. 
These two words share the same gesture, but differ in the non-manual features. 
The latter is discriminative in the movements of the head and eyes, which only cover the small spatial size of the whole image. 
It can be observed that the saliency map is able to adaptively locate the discriminative cues. 
The left part concentrates on the hand region, while the right part focuses on not only the hand region but also the facial region. 
Besides, these densely sampled areas vary along the time.
With these adaptive sampled details, the rest of convolutional stages are easier to focus on the discriminative fine-grained cues.

\textbf{Mutual promotion module.} 
Extensive experiments are conducted to show the complementary effects of the two modules mentioned above and validate the effectiveness of our mutual promotion module. 
\begin{table}
\small
\begin{center}
\tabcolsep=8pt
\begin{tabular}{cccccccc}
\hline
Backbone & GE         & LE         & COM        & MP & Top-1 & Top-2 & Top-5   \\ \hline  \hline
3D-R50   &            &            &            &    & 62.1 & 73.2  & 82.9  \\
3D-R50   & \checkmark &            &            &    & 64.8 & 76.1  & 84.7  \\
3D-R50   &            & \checkmark &            &    & 63.9 & 73.8  & 83.0  \\
3D-R50   & \checkmark & \checkmark &            &    & 66.9 & 78.2  & 86.2  \\
3D-R50   & \checkmark & \checkmark & \checkmark &    & 67.4 & 78.0  & 86.6  \\
3D-R50   & \checkmark & \checkmark & \checkmark & \checkmark & \textbf{69.9} & \textbf{82.2} & \textbf{90.5} \\ \hline 
\end{tabular}
\end{center}
\caption{Ablation study on the effects of each loss term. GE, LE and COM denotes the loss supervising the global enhancement stream, local enhancement stream and their fusion prediction, respectively, while MP stands for the mutual promotion loss. 3D-R50 denotes 3D-ResNet-50 backbone. (Top-1, Top-2 and Top-5 accuracy (\%)).}
\label{mutual}
\end{table}	
During inference, we fuse the prediction results from the global and local stream.
As shown in Table~\ref{mutual}, by combining the two streams with a simple sum fusion, the top-1 accuracy is boosted from 62.1\% to 66.9\%, where the performance gain is larger than the sum of what a single enhancement module brings.
With $\mathcal{L}_{com}$ added, the top-1 accuracy improves from 66.9\% to 67.4\%.
These results demonstrate the complementary effect of these two enhancement modules.
One focuses on the global contextual motion cues, while the other targets on the discriminative cues hidden in the fine-grained details. 

Besides the summation fusion as a quite simple fusion strategy, we integrate the mutual promotion module to further enhance the performance of the whole framework. 
With this module, the top-1 accuracy is further improved to 69.9\%. By referring to each other, each stream is able to modify its own learning process. 
By using the local stream as reference, the global stream can learn from the knowledge fine-grained cues contain. 
Hence, it will help produce better feature maps to guide the sampling process and improve the local stream. 
Through this cycling way, both streams will benefit from each other, thus the accuracy of the whole network gets improved.

\subsection{Results and Analysis}
\textbf{Evaluation on NMFs-CSL dataset.} We compare our approach with other challenging methods on NMFs-CSL dataset. 
To further show the challenges of our proposed dataset, we divide the \emph{total} test set into two subsets containing the aforementioned non-manual-features-aware \emph{confusing} words and the \emph{normal} words, respectively.
We also report the accuracy on them.
From Table~\ref{full}, the accuracy on the confusing set lays largely behind the normal set, which validates the challenge introduced by the NMFs-aware confusing words.
It can be observed that 3D-CNN outperforms 2D-CNN by a large margin, which validates the high capacity of 3D-CNN for modeling spatio-temporal representation. 
DNF~\cite{cui2019deep} achieves challenging performance in continuous SLR. 
We extract its visual encoder as the classification backbone for this task.
It achieves 55.8\% top-1 accuracy on the total set, which is inferior to the 3D-R50 baseline.
This is due to the fact that isolated and continuous SLR orient at different aspects, \emph{i.e.,} word-level and sentence-level recognition, respectively.
Then we compare with the methods of general video classification, such as I3D~\cite{carreira2017quo}, Slowfast~\cite{feichtenhofer2019slowfast}, \emph{etc.}
I3D~\cite{carreira2017quo} is a challenging method among them. 
It obtains 64.4\%, 47.3\% and 87.1\% top-1 accuracy on the total, confusing and normal set, respectively.
Slowfast~\cite{feichtenhofer2019slowfast} is a state-of-the-art method in general video classification, which capitalizes two paths at different frame rates for temporal and spatial semantics, respectively.
It reaches 66.3\%, 47.0\% and 92.0\% top-1 accuracy on the total, confusing and normal set, respectively.

Even compared with these challenging methods, our method still achieves state-of-the-art performance on three reported sets.
It is notable that the performance gain introduced by our methods is larger on the confusing set than that on the normal set.
Specifically, our method achieves 4.8\% and 2.1\% top-1 accuracy gain over the most challenging Slowfast~\cite{feichtenhofer2019slowfast} method on the confusing and normal set, respectively.
Similar results are also observed when comparing with other general video classification methods.
These general methods may not be suitable for capturing the fine-grained cues hidden in the sign video.

\begin{table*}
\small
\begin{center}
\begin{tabular}{l|ccc|ccc|ccc}
\hline
\multirow{2}{*}{Method}      &  \multicolumn{3}{c|}{Total} &  \multicolumn{3}{c|}{Confusing} &  \multicolumn{3}{c}{Normal}  \\ 
         & Top-1 & Top-2 & Top-5 & Top-1 & Top-2 & Top-5 & Top-1 & Top-2 & Top-5 \\ \hline \hline 
2D-R50~\cite{wang2016temporal} & 56.3  & 71.3  & 83.7 & 33.2 & 54.2 & 73.3 & 87.1 & 94.2 & 97.7 \\
3D-R50~\cite{Qiu_2017_ICCV}    & 62.1 & 73.2 & 82.9 & 43.1 & 57.9 & 72.4 & 87.4 & 93.4 & 97.0 \\
DNF~\cite{cui2019deep}         & 55.8 & 69.5 & 82.4 & 33.1 & 51.9 & 71.4 & 86.3 & 93.1 & 97.0 \\
MF-Net~\cite{chen2018multi}    & 64.3 & 78.7 & 88.6 & 44.5 & 65.6 & 81.2 & 90.7 & 96.3 & 98.9 \\
I3D~\cite{carreira2017quo}     & 64.4 & 77.9 & 88.0 & 47.3 & 65.7 & 81.8 & 87.1 & 94.3 & 97.3 \\
TSM~\cite{lin2019tsm}          & 64.5 & 79.5 & 88.7 & 42.9 & 66.0 & 81.0 & 93.3 & 97.5 & 99.0 \\
Slowfast~\cite{feichtenhofer2019slowfast} & 66.3 & 77.8 & 86.6 & 47.0 & 63.7 & 77.4 & 92.0 & 96.7 & 98.9 \\ \hline \hline 
Ours                           & \textbf{69.9} & \textbf{82.2} & \textbf{90.5} & \textbf{51.8} & \textbf{70.6} & \textbf{83.9} & \textbf{94.1} & \textbf{97.8} & \textbf{99.4} \\ \hline
\end{tabular}
\end{center}
\caption{Accuracy comparison on NMFs-CSL dataset. `Total', `Confusing' and `Normal' denote the accuracy over all the words, the NMFs-aware confusing words, and the rest of words in the test set, respectively. (Top-1, Top-2 and Top-5 accuracy (\%)).}
\label{full}
\vspace{-0.5cm}
\end{table*}

\begin{figure*}[ht]
\centering
\subfigure[Group1: 3D-ResNet-50 Baseline]{
\label{a}
\begin{minipage}[b]{0.4\linewidth}
\centering
\includegraphics[width=0.8\linewidth]{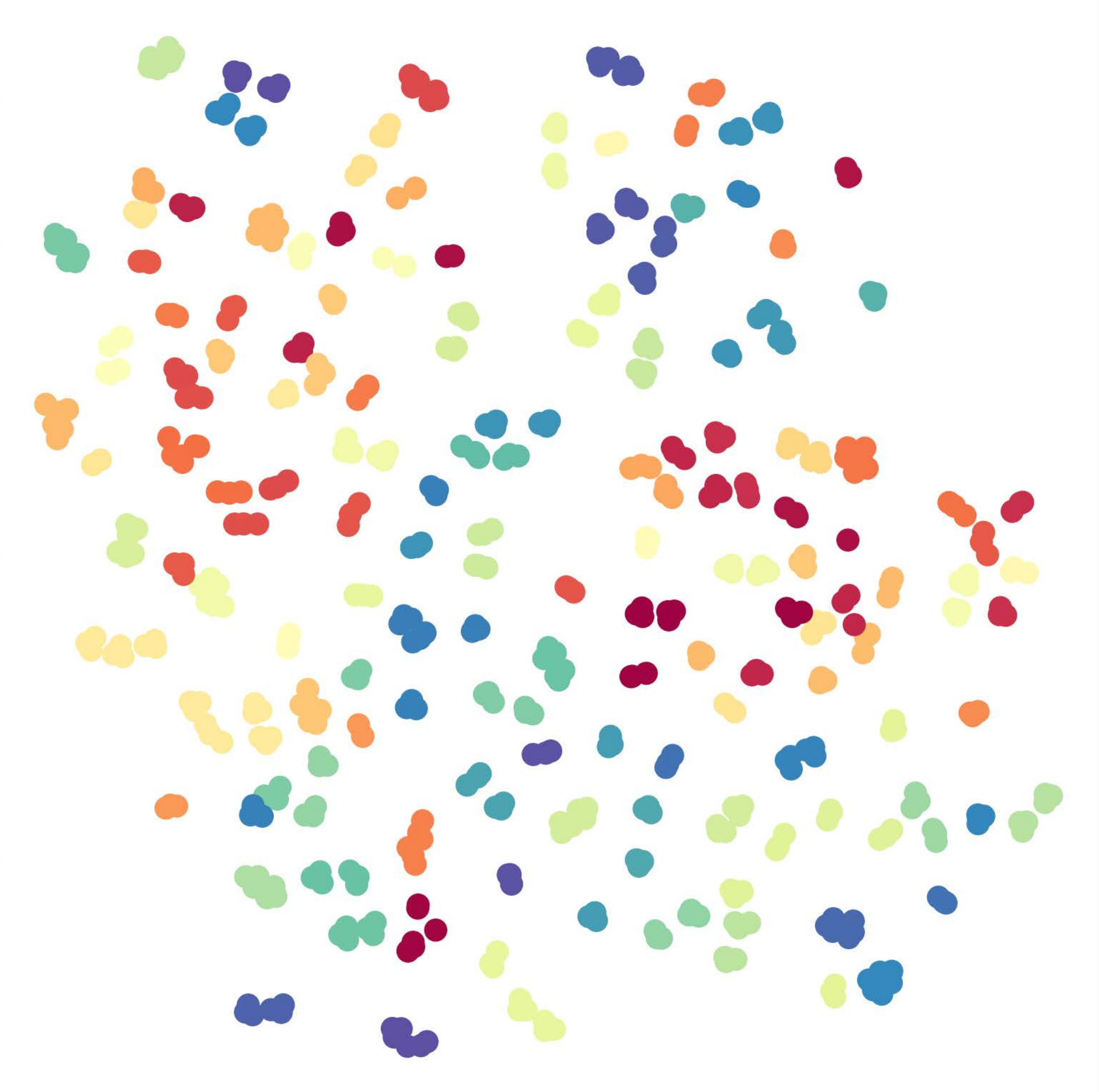}\vspace{4pt}\end{minipage}}
\subfigure[Group1: Our GLE-Net]{
\label{b}
\begin{minipage}[b]{0.4\linewidth}
\centering
\includegraphics[width=0.8\linewidth]{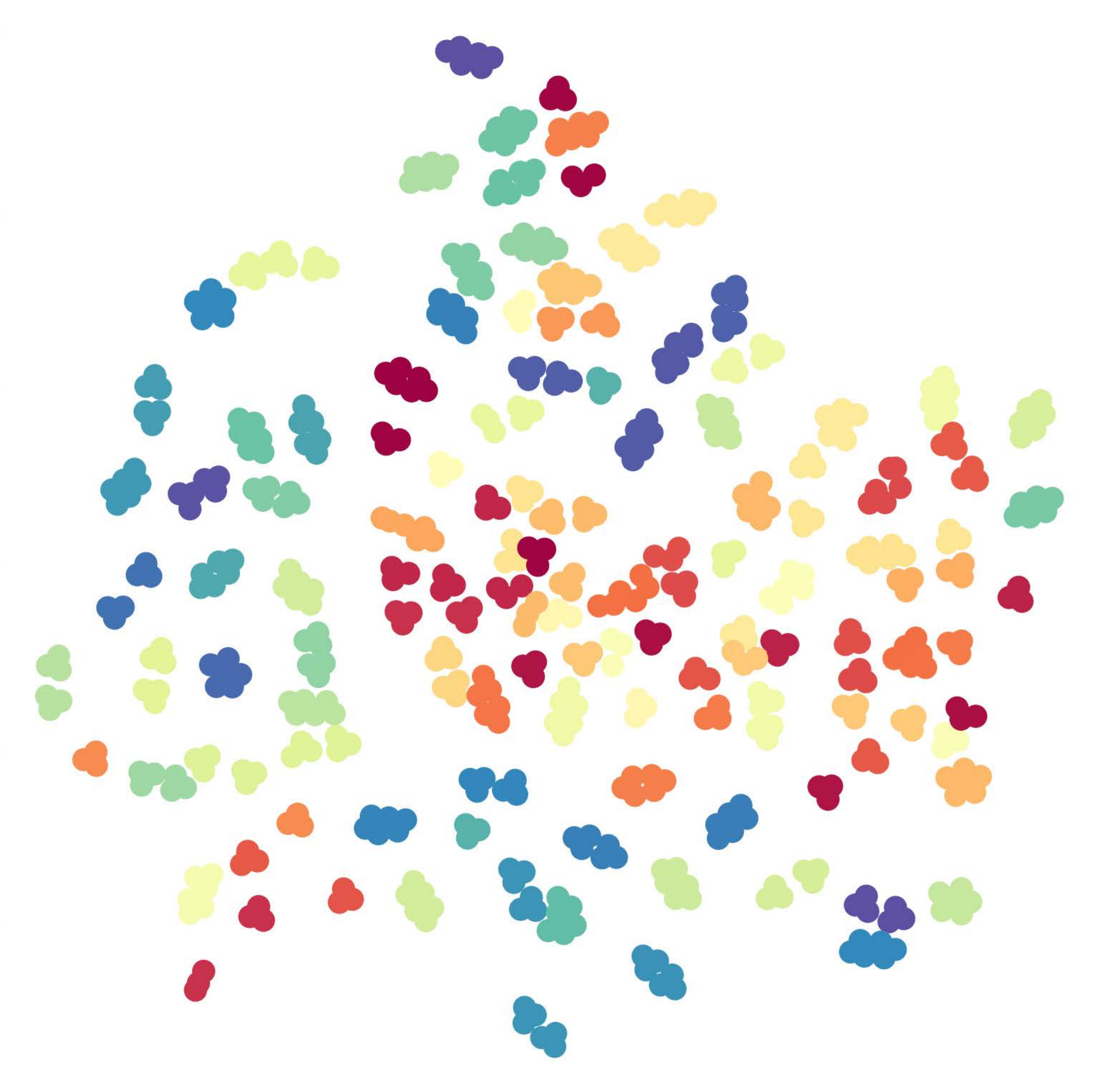}\vspace{4pt}\end{minipage}}
\centering
\subfigure[Group2: 3D-ResNet-50 Baseline]{
\label{c}
\begin{minipage}[b]{0.4\linewidth}
\centering
\includegraphics[width=0.8\linewidth]{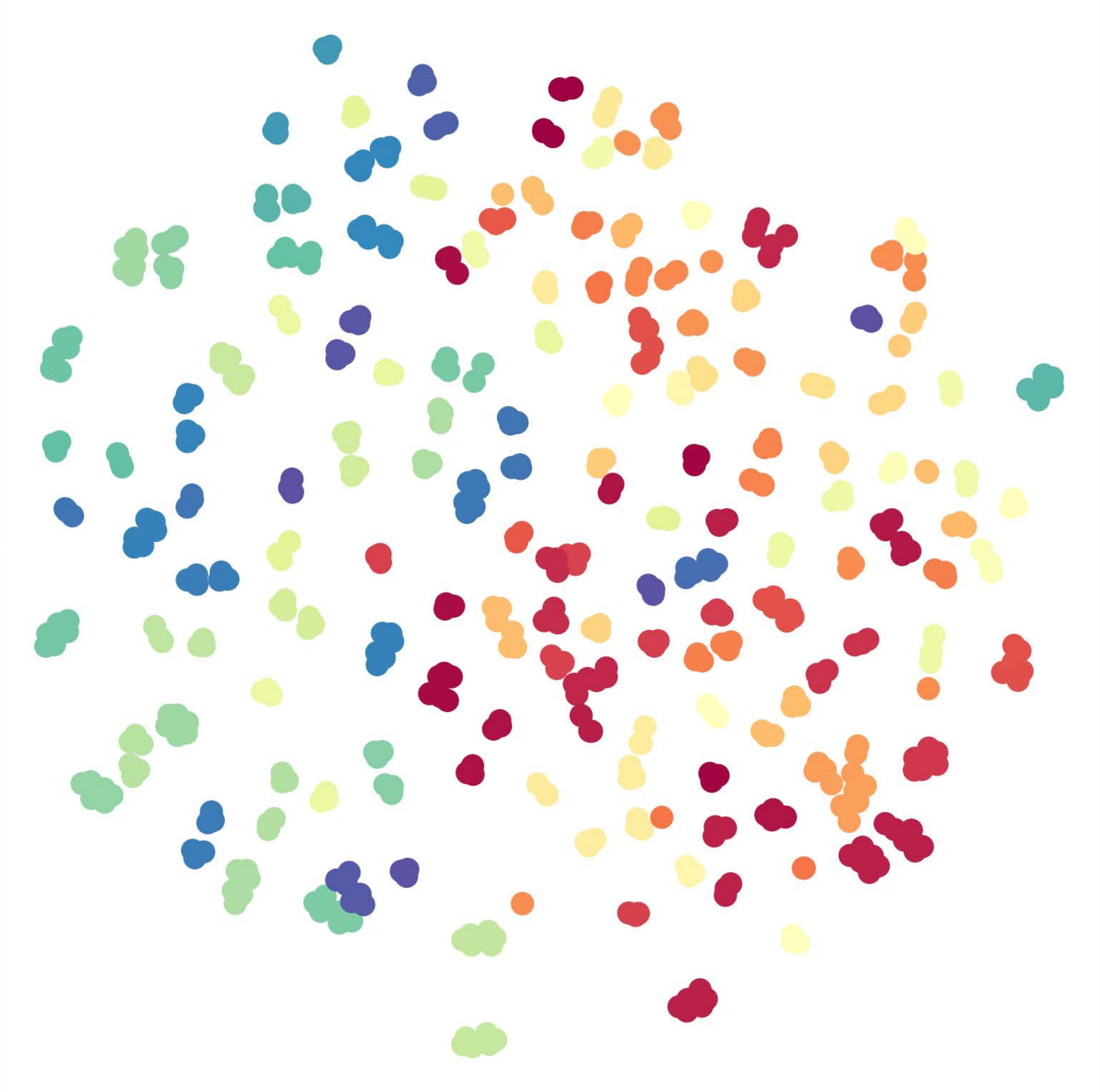}\vspace{4pt}\end{minipage}}
\subfigure[Group2: Our GLE-Net]{
\label{d}
\begin{minipage}[b]{0.4\linewidth}
\centering
\includegraphics[width=0.8\linewidth]{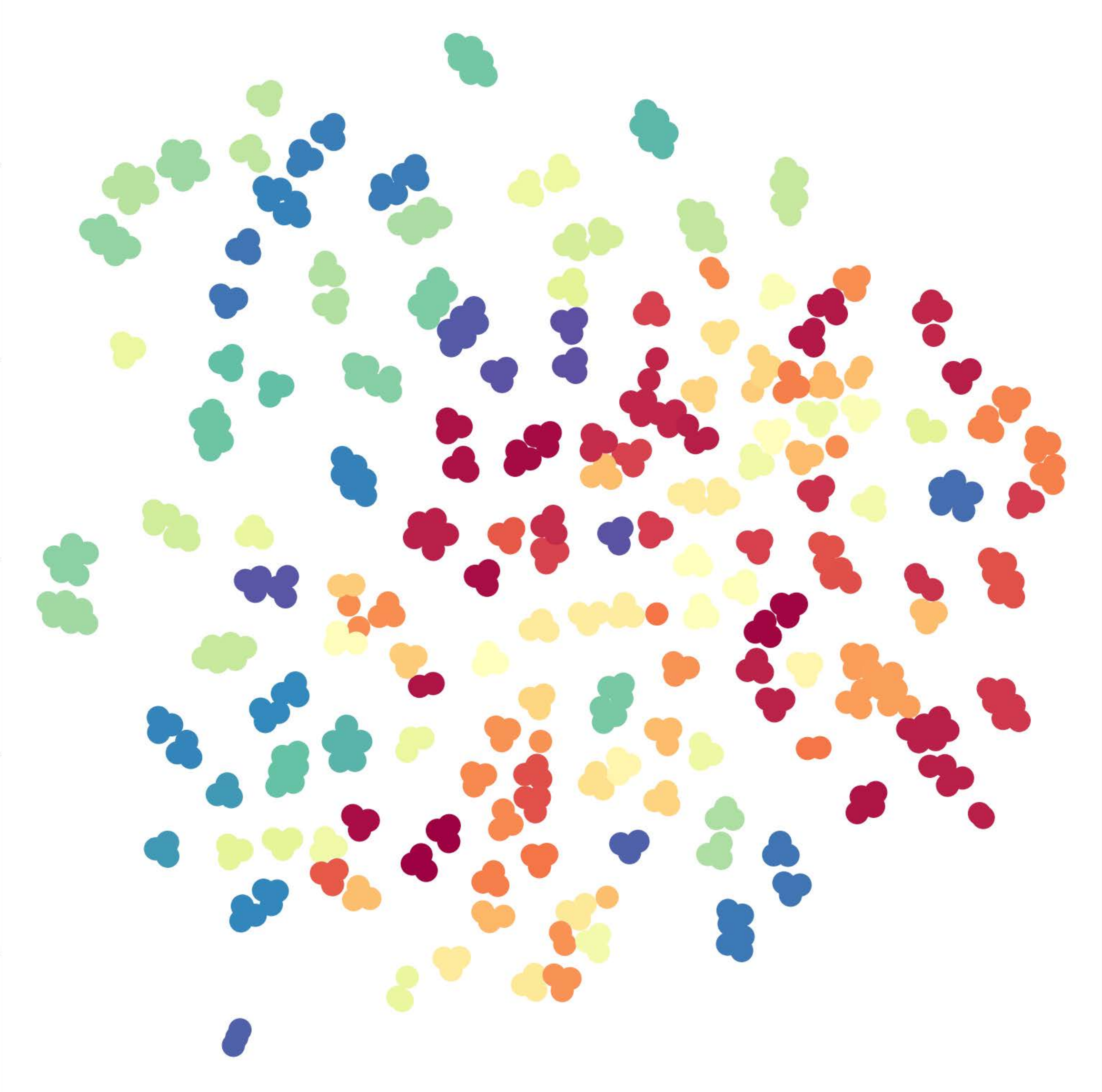}\vspace{4pt}\end{minipage}}
\caption{Feature embedding visualization of the 3D-ResNet-50 baseline and our method on NMFs-CSL dataset using t-SNE~\cite{maaten2008visualizing}. We randomly select 100 classes two times (denoted as Group1 and Group2) on the test set. Each sample is visualized as a point. Samples belonging to the same class have the same color.}
\label{t-sne}
\end{figure*}

To further illustrate the feature representations enhanced by our proposed methods, we visualize the feature embeddings using t-SNE~\cite{maaten2008visualizing}.
For accurate recognition, samples belonging to the same class are expected to cluster together during visualization.
We randomly select 100 classes two times, and visualize these two groups in two rows. 
Samples in the testing set are visualized, \emph{i.e.,} 6 samples for each class.
Each sample is visualized as a point.
Notably, samples belonging to the same class are visualized as the same color, \emph{i.e.,} different colors indicates different classes.
We compare the feature embeddings produced by the 3D-ResNet-50 baseline and our method.
It can be observed that the features generated by the baseline are distributed randomly over a wide area and the samples belonging to the same class cannot gather together. 
In contrast, for our method, the points belonging to the same class are clustered more tightly, while points from different classes are mutually exclusive. 
These qualitative results demonstrate the effectiveness of our method.

\begin{table}
\small
\begin{center}
\tabcolsep=6pt
\begin{tabular}{l|cccc|c}
\hline
\multirow{2}{*}{Method}      &  \multicolumn{4}{c|}{Modality} &  \multirow{2}{*}{Accuracy}   \\ 
         & RGB & OF & Depth & Skeleton &  \\ \hline \hline 
STIP~\cite{laptev2005space}    & \checkmark  &            &            &            &  61.8 \\
iDT~\cite{wang2011action}     & \checkmark  & \checkmark &            &            &  68.5 \\
GMM-HMM~\cite{tang2015real} & \checkmark  &            & \checkmark & \checkmark &  56.3 \\
C3D~\cite{tran2015learning}     & \checkmark  &            & \checkmark &            &  74.7 \\
Atten~\cite{huang2018attention}   & \checkmark  &            & \checkmark & \checkmark &  88.7 \\ \hline \hline
Ours    & \checkmark  &            &            &            & \textbf{96.8}      \\ \hline
\end{tabular}
\end{center}
\caption{Accuracy comparison on SLR500 dataset. (Top-1 accuracy (\%)).}
\label{slr500}
\end{table}

\textbf{Evaluation on SLR500 dataset.} We also perform experiments on SLR500~\cite{huang2018attention} dataset, as shown in Table~\ref{slr500}.
Besides the RGB modality, this dataset also provides depth and skeleton data. 
Atten~\cite{huang2018attention} achieves promising performance on this dataset, which designs attention mechanism on the spatial and temporal dimension and utilizes three modalities for inference.
Our method only uses RGB data as input and achieves a significant performance improvement over Atten, with a 8.1\% top-1 accuracy gain.
It reveals that our method not only distinguishes these NMFs-aware confusing sign words, but also improves the performance on the normal words.

\section{Conclusion}
\label{sec:conclusion}
In this paper, we are dedicated to the recognition of confusing isolated sign language where non-manual features play an important role. 
The NMFs-aware confusing words are visually similar and only differ in non-manual features, \emph{e.g.} facial expressions, eye gaze, \emph{etc.}, which introduce great challenges in accurate SLR.
To tackle the above issue, we propose a Global-local Enhancement Network~(GLE-Net). 
It aims to enhance feature representations from two complementary aspects, \emph{i.e.,} the global enhancement module for contextual information, and the local enhancement module for fine-grained cues. 
These two modules are optimized in mutual promotion for further performance improvement. 
For the validation of our method, we introduce a non-manual-features-aware isolated Chinese sign language dataset~(NMFs-CSL).
To our best knowledge, it is the first benchmark dataset explicitly emphasizing the importance of NMFs. 
Extensive experiments validate the effectiveness of our method not only for these confusing words, but also for normal words in sign language recognition.
Our method achieves state-of-the-art performance on two benchmark datasets.
We hope the proposed NMFs-CSL dataset and GLE-Net will further promote SLR research.

\begin{acks}
This work was supported by NSFC under contract No. 61632019, 61836006 and 61836011, and Youth Innovation Promotion Association CAS (No. 2018497).
It was also supported by the GPU cluster built by MCC Lab of Information Science and Technology Institution, USTC.
\end{acks}

\bibliographystyle{ACM-Reference-Format}
\bibliography{ref}
\end{document}